%% file: main-arxiv20230308.tex
\documentclass{article}

\usepackage[english]{babel}
\usepackage{iclr2023,times}

\usepackage[letterpaper,top=2cm,bottom=2cm,left=3cm,right=3cm,marginparwidth=1.75cm]{geometry}
\usepackage{wrapfig}
\usepackage{booktabs}

\usepackage{amsmath}
\usepackage{amsfonts}
\usepackage{amssymb}
\usepackage{mathrsfs}
\usepackage{algorithm}
\usepackage{algorithmic}
\usepackage{braket}
\usepackage{graphicx}
\usepackage[]{pifont}
\usepackage{xcolor}
\usepackage[colorlinks=true, allcolors=blue]{hyperref}

\newcommand{\cA}{\mathcal{A}}
\newcommand{\cD}{\mathcal{D}}

\newcommand{\cS}{\mathcal{S}}
\newcommand{\cX}{\mathcal{X}}
\newcommand{\cY}{\mathcal{Y}}
\newcommand{\cZ}{\mathcal{Z}}

\newcommand{\RR}{\mathbb{R}}

\DeclareMathOperator*{\argmin}{argmin}
\DeclareMathOperator*{\BR}{BR}

\title{Mastering Strategy Card Game (Legends of Code and Magic) via End-to-End Policy and Optimistic Smooth Fictitious Play}

\author{Wei Xi \\
{\normalsize \ ByteDance} \\
{\tiny\texttt{weatwe1999@gmail.com }} \\
\And
Yongxin Zhang \\
{\normalsize \ ByteDance} \\
{\tiny\texttt{zhangyongxin.yx@bytedance.com}} \\
\And
Changnan Xiao \\
{\normalsize \ ByteDance} \\
{\tiny\texttt{changnanxiao@gmail.com}} \\
\And
Xuefeng Huang \\
{\normalsize \ ByteDance} \\
{\tiny\texttt{wangfuming@bytedance.com}} \\
\And
Shihong Deng \\
{\normalsize \ ByteDance} \\
{\tiny\texttt{dengshihong@bytedance.com}} \\
\And
Haowei Liang \\
{\normalsize \ ByteDance} \\
{\tiny\texttt{lianghaowei@bytedance.com}} \\
\And
Jie Chen \\
{\normalsize \ ByteDance} \\
{\tiny\texttt{chenjiexjtu@gmail.com}} \\
\And
Peng Sun\thanks{Corresponding author.} \\
{\normalsize \ ByteDance} \\
{\tiny\texttt{pengsun000@gamil.com}} 
}

\iclrfinalcopy % Uncomment for camera-ready version, but NOT for submission.

\begin{document}

\maketitle

\begin{abstract}
Deep Reinforcement Learning combined with Fictitious Play shows impressive results on many benchmark games, 
most of which are, however, single-stage.
In contrast, 
real-world decision making problems may consist of multiple stages, 
where the observation spaces and the action spaces can be completely different across stages. 
We study a two-stage strategy card game \emph{Legends of Code and Magic} and propose an end-to-end policy to address the difficulties that arise in multi-stage game. 
We also propose an optimistic smooth fictitious play algorithm to find the Nash Equilibrium for the two-player game. 
Our approach wins double championships of COG2022 competition. 
Extensive studies verify and show the advancement of our approach. 
\end{abstract}

\section{Introduction}
\label{sec:intro}

Deep Reinforcement Learning has shown many impressive results on different kinds of single- or multi- player games. 
\citep{silver2016mastering} achieves superhuman performance on the two-player board game Go.
\citep{berner2019dota} defeats the world champion team on the 5-vs-5 MOBA game Dota2. 
\citep{vinyals2019grandmaster} reaches grandmaster level on the two-player RTS game StarCraft2. 
\citep{badia2020agent57} achieves human level on all 57 Atari games, which contains different kinds of single-player video games. 
\citep{li2020suphx} reaches the highest level of the four-player board game Mahjong on Tenhou platform. 

However, these breakthroughs concentrate on one-stage games, where the observation space and the action space are kept unchanged during the whole gameplay. 
For instance, the two-stage 5-vs-5 MOBA game Dota2 contains a prerequisite ban-pick stage, where players must choose heroes that are used in the later battle stage. 
But the policy of ban-pick stage is decided by rules/expert experiences \citep{berner2019dota} or a separate searching based method \citep{ye2020towards}, 
which simplifies Dota2 to a one-stage game. 
It's never uncommon that a game or a real world decision making problem contains multiple stages with completely different observation spaces and action spaces. 

Strategy card game is a well-known genre that contains two nontrivial stages, 
which are card-deck building stage and battle stage. 
In card-deck building stage, the player selects cards from a card pool to construct a deck. 
In battle stage, the player plays cards that are randomly drawn from the deck to fight against the opponent. 
In this paper, we study a strategy card game, \emph{Legends of Code and Magic} \citep{locm}, and propose an end-to-end policy function to tackle the multi-stage problem. 

Most strategy card games are two-player zero-sum games, 
where Nash Equilibrium is usually viewed as the optimal policy.
We propose an \emph{Optimistic Smooth Fictitious Play} algorithm that efficiently finds Nash Equilibrium for large-scale multi-step game with last-iterate convergence, 
and is conveniently implemented with Deep Reinforcement Learning as sub-solver combined with opponent model sampling.
 
Our main contributions are as follow:
\begin{itemize}
    \item We propose an optimistic smooth fictitious play algorithm to solve the two-player zero-sum game, where we apply deep reinforcement learning as the smooth best response solver. 
    \item We propose an end-to-end policy function that addresses the multi-stage game. 
    \item We verify and show advancement of our approach in the strategy card game Legends of Code and Magic. 
\end{itemize}

\begin{figure}[h]
\centering
\includegraphics[width=1\textwidth]{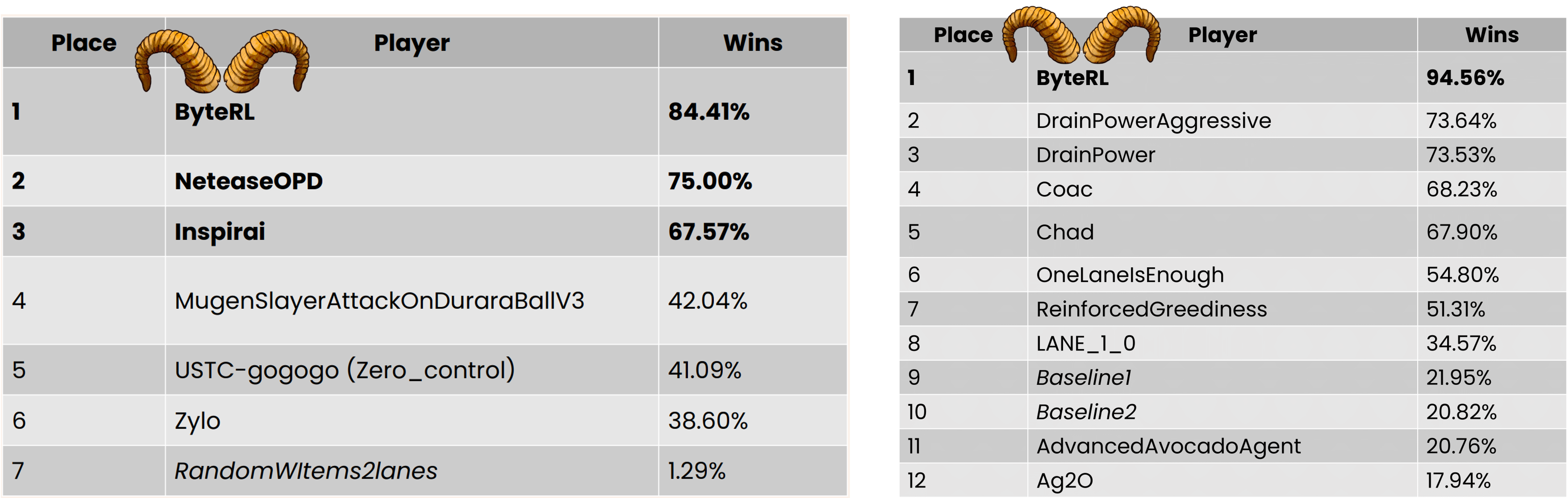}
\caption{COG2022 LoCM 1.5 competition (left) and LoCM 1.2 competition (right). Our approach achieves the 1st place in both tracks.}
\label{fig:reuslt}
\end{figure}

\section{Method}

% develop OSFP in Normal Form Game
\subsection{Optimistic Smooth Fictitious Play}
\label{sec:osfp}

% 2p0s game, bilinear matrix, NE
Consider a finite \emph{two-player zero-sum} (2p0s) game. 
Player 1 policy $x \in \cX$ and player 2 policy $y \in \cY$ are probabilities over the respective feasible actions.
The payoff function $u(x, y)$, 
which player 1 wants to minimize and player 2 wants to maximize,
is bi-linear
\begin{equation}
    u(x,y) = x^{\top} A y, \quad x \in \cX, y \in \cY
    \label{eq:utility}
\end{equation}
with the payoff matrix $A \in \RR^{K \times K}$.
In single step simultaneous-action game (or \emph{Normal Form Game}), 
player 1 policy space $\cX$ is a \emph{simplex} $\Delta^K$ where $K$ is the number of feasible actions that player 1 can take,
e.g., in the Rock-Paper-Scissors we have $K=3$ and $x \in \cX$ is a $3$-dimensional vector satisfying non-negativity and sum-to-one.
While in multi-step turn-based game, 
$\cX$ is a \emph{Treeplex} \citep{kroer2017theoretical,hoda2010smoothing,gilpin2007lossless},
where $x$ is an ultra long $K$-dimensional vector with each component being the \emph{traverse-probability} through a player 1's edge in the \emph{Game Tree} so as to satisfy the incoming-outgoing probability mass constraint.
An example is the strategy card game as considered in this paper.
The above discussion goes similar for player 2 policy $y \in \cY$.

% The Nash Equilibrium (NE)
The policy $(x^*, y^*)$ is called \emph{Nash Equilibrium} (NE) if it satisfies that $u(x^*, y) \le u(x^*,y^*) \le u(x,y^*)$. 
In 2p0s game, NE can be viewed as an optimal policy in the sense that neither player can improve his payoff by unilaterally deviating from the NE policy.

% the VI form.
For a compact expression and for convenience of the following discussion, 
we re-formulate the 2p0s game in \emph{Variational Inequality} (VI) form \citep{facchinei2003finite,mertikopoulos2019learning}. 
Let $z=(x,y)^{\top} \in \cZ = \cX \times \cY$. 
Define the \emph{payoff vector} $F(z)$ as the concatenation of the \emph{individual gradient} as
\begin{equation}
    F(z) = (\nabla_x u(x,y), -\nabla_y u(x,y))^{\top} = (Ay, -A^{\top}x)^{\top} \equiv (F(y), F(x))^{\top}.
    \label{eq:payoff-vector}
\end{equation}
Note that $F(z)$ itself is not necessarily a gradient of some scalar function (Sec.1.3.1, Sec.1.4.2 of \citep{facchinei2003finite}) and should be understood as a vector field over domain $\cZ$.

% Best Response (BR) in VI, say it splits into two player's seperate BRs.
For a policy $z \in \cZ$, the \emph{Best Response} (BR) to this policy is defined as
\begin{equation}
    \BR(z) = \argmin_{z' \in \cZ} \braket{F(z), z'} = (\argmin_{x' \in \cX} \braket{Ay, x'}, \argmin_{y' \in \cY} \braket{-A^{\top}x, y'})^{\top},
    \label{eq:vi-br}
\end{equation}
which can be understood as the concatenation of player 1's and 2's individual BRs.

% NE in VI
An equivalent definition of NE is that the policy $z^*=(x^*,y^*)$ is the best response to itself, 
i.e., $z^*$ is a \emph{fixed point} of the best response operator:
\begin{equation*}
    z^* = \BR(z^*) = \argmin_{z \in \cZ} \braket{F(z^*), z}.
\end{equation*}
The NE condition and BR definition imply that $\braket{F(z^*),z^*} \le \braket{F(z^*),z}$.
Noting this, we can also write NE in VI condition:
\begin{equation*}
    \braket{F(z^*), z^*-z} \le 0, \quad z \in \cZ.
    \label{eq:vi-ne}
\end{equation*}

% say that BR can be multi-value, introduce smooth BR in VI, define \psi in VI. 
The $\BR(\cdot)$ as in Eq. \eqref{eq:vi-br} is in general a set-valued function since there can be multiple $z' \in \cZ$ that minimizes $\braket{F(z),z'}$.
To make the BR policy unique, 
\emph{Smooth Best Response} (SBR) policy is introduced in the literature \citep{fudenberg1998theory},
whose VI form is:
\begin{equation}
    \BR\hspace{+0.0001em}_{\psi}(z) = \argmin_{z' \in \cZ} \{ \Braket{F(z), z'} + \psi(z') \},
\label{eq:vi-sbr}
\end{equation}
where $\psi(\cdot)$ is a \emph{regularizer} defined on domain $\cD \subseteq \cZ$.
Usually, $\psi(\cdot)$ is chosen to be a strongly convex function that is well-defined on domain $\cZ$. 
For instance, in single step game, $\psi(\cdot)$ can be the sum of player 1's and 2's \emph{negative entropy} 
\begin{equation*}
    \psi(z)=\psi_1(x) + \psi_2(y)=\sum_{i=1}^K x_i \log(x_i) + \sum_{i=1}^K y_i \log(y_i),
\end{equation*}
where $\cD = \Delta^K \times \Delta^K$.
Also, \citep{kroer2017theoretical,hoda2010smoothing,gilpin2007lossless} have extended the entropy function $\psi(\cdot)$ to be defined on Treeplex for multi-step game.

As the payoff vector $F(z) = (Ay, -A^{\top}x)^{\top}$ is solely determined by the policy $z=(x,y)^{\top}$,
we can overload the notations BR Eq. \eqref{eq:vi-br} and SBR Eq. \eqref{eq:vi-sbr} as:
\begin{equation}
    \BR(F(z)) \equiv \BR(z), \quad\quad  \BR\hspace{+0.0001em}_{\psi}(F(z)) \equiv \BR\hspace{+0.0001em}_{\psi}(z),
\end{equation}
which emphasize that the (smooth) best response can be calculated by observing only the payoff vector $F(z)$.

% Give the Optimistic Smooth Fictitious Play (OSFP), say it uses smooth BR as building block
To this extent,
we want an efficient NE finding algorithm.
For this purpose,
we describe the \emph{Optimistic Smooth Fictitious Play} (OSFP), 
which adopts SBR as building block to find NE in an iterative algorithm.
Specifically, at iteration $k$ it updates the policy $z$ by taking the SBR to the mixture of the historical policies:
\begin{equation}
    z_{k+1} = \BR\hspace{+0.0001em}_{\psi}(\eta(\tilde{F}_k+F_k)) = \argmin_{z \in \cZ} \left \{ \eta\braket{\tilde{F}_k + F_k, z} + \psi(z) \right \},
    \label{eq:osfp}
\end{equation}
where $\eta$ is some step-size and the subscript $k$ indicates ``at iteration $k$''.
The abbreviated notation $F_k$ denotes the payoff vector $F_k \equiv F(z_k)$, 
where $z_k=(x_k,y_k)$ denotes the policy at iteration $k$.
Also, denote by 
\begin{equation}
    \tilde{F}_k \equiv \sum_{t=1}^k F_t = \left ( A \left ( \sum_{t=1}^k y_t \right ), -A^{\top} \left (\sum_{t=1}^k x_t \right ) \right )^{\top} 
\end{equation}
the mixture of historical payoff vectors (or historical policies).
Note that in Eq. \eqref{eq:osfp} the payoff vector $F_k$ is ``taken twice'' when calculating the SBR.
The ``extra'' $F_k$ should be interpreted as the optimistic prediction on the opponent's future policy \citep{rakhlin2013online}, 
henceforth the name OFSP.
If we drop the optimistic term, then Eq. \eqref{eq:osfp} reduces to \emph{Smooth Fictitious Play} (SFP):
\begin{equation}
    z_{k+1} = \BR\hspace{+0.0001em}_{\psi}(\eta(\tilde{F}_k)) = \argmin_{z \in \cZ} \{ \eta\braket{\tilde{F}_k , z} + \psi(z) \},
    \label{eq:sfp}
\end{equation}

% Briefly discuss OSFP ans SFP. SFP: average sequence converge, but actual sequence can cycle; OSFP: actual sequence converges. 
The SFP \eqref{eq:sfp} is also known as \emph{Follow-The-Regurlarized-Leader} (FTRL) in the online learning literature \citep{mertikopoulos2019learning,shalev2012online} or Dual Averaging in the mathematical programming literature \citep{nesterov2009primal}.
FTRL is shown to be \emph{no-regret}, therefore by a ``folk theorem'' (e.g., see \citep{cesa2006prediction}) the \emph{average-sequence} $\{ \Bar{z}_k \}_{k \ge 1}$ converges to NE, where $\Bar{z}_k \equiv \frac{1}{k} \sum_{t=1}^k z_k$.
Unfortunately, the \emph{actual-sequence} $\{ z_k \}_{k \ge 1}$ of FTRL/FSP does not converge and can cycle around NE \citep{mertikopoulos2018cycles}.
In contrast, we can show that the actual sequence of OFSP converges to NE, 
a property that is referred to as \emph{last-iterate} convergence in the literature \citep{syrgkanis2015fast,daskalakis2018last,wei2021linear}. 
% Briefly discuss that OSFP is convinient for RL impl, while SFP is not due to the averaging?
The last-iterate convergence is more convenient, 
as we only need maintain the policy being trained and do not need a separate process to learn the average policy. 
This is especially friendly when we apply it to multi-step game and use reinforcement learning as sub-solver as discussed in Sec. \ref{sec:drl}. 

% Discuss that why OSFP converges. citep Luo Haipeng paper, show how OSFP is equivalent to Optimisti Mirror Descent (OMD). 
\citep{wei2021linear} propose the \emph{Optimistic Mirror Descent} (OMD) algorithm,
which is shown to find NE in a linear convergence rate.
Actually, using some standard techniques (see, e.g., Sec.28 of \citep{lattimore2020bandit}),  
we can show that OFSP is equivalent to OMD with a different implementation when $\cD$, 
the domain of the regularizer $\psi(\cdot)$, 
satisfies that $\cD \subseteq \cZ$, 
which is explained as follows. 

In \citep{wei2021linear}, 
the OMD algorithm updates the policy as
\begin{align}
    z_k   &= \argmin_{z \in \cZ} \{ \eta \braket{F_{k-1}, z} + D_{\psi}(z, \hat{z}_k) \} \label{eq:omd1} \\
    \hat{z}_{k} &= \argmin_{z \in \cZ} \{ \eta \braket{F_{k-1}, z} + D_{\psi}(z, \hat{z}_{k-1}) \}, \label{eq:omd2}
\end{align}
where $D_{\psi}$ is \emph{Bregman Divergence}:
\begin{equation*}
    D_{\psi}(z, z') = \psi(z) - \psi(z') - \braket{\nabla \psi(z'), z - z'}.
\end{equation*}
In Eq. \eqref{eq:omd1} and \eqref{eq:omd2},
$\{ z_k \}_{k \ge 1}$ is the primary sequence that converges to NE, 
while $\{ \hat{z}_k \}$ is a secondary (auxiliary) sequence for the convenience of computation.

Noting $\cD \subseteq \cZ$, 
the first-order optimal condition of Eq. \eqref{eq:omd1} or \eqref{eq:omd2} can be obtained by setting to zero the gradient,
and the stationary point is guaranteed to reside in $\cZ$. 
It thus yields
\begin{align}
    \eta F_{k-1} + \nabla \psi(z_k) -\nabla \psi(\hat{z}_k) &= 0, \label{eq:omd-grad1} \\
    \eta F_{k-1} + \nabla \psi(\hat{z}_k) - \nabla \psi(\hat{z}_{k-1}) &= 0, \label{eq:omd-grad2}
\end{align}
where Eq. \eqref{eq:omd-grad1}, \eqref{eq:omd-grad2} are derived from Eq. \eqref{eq:omd1}, \eqref{eq:omd2}, respectively.
Summing up Eq. \eqref{eq:omd-grad1}, \eqref{eq:omd-grad2} and telescoping Eq. \eqref{eq:omd-grad2} through $k,k-1,...,2,1$ with the initial value $\hat{z}_1$ satisfying $\nabla\psi(\hat{z}_1)=0$,
we have
\begin{equation*}
    \eta F_{k-1} + \eta \sum_{t=1}^{k-1} F_t + \nabla\psi(z_k)=0,
\end{equation*}
which gives
\begin{equation}
    z_k = \nabla \psi^{-1}(-\eta F_{k-1} - \eta \tilde{F}_{t-1}). 
    \label{eq:omd-solution}
\end{equation}
Note that Eq. \eqref{eq:omd-solution} happens to be the solution to OSFP \eqref{eq:osfp},
which is immediately verified by setting to zero the gradient of \eqref{eq:osfp} for iteration $k$.

Therefore, the convergence and the speed of OMD \citep{wei2021linear} carries over to OSFP.

% show how OSFP is implemented with DRL
\subsection{Solving SBR via Deep Reinforcement Learning}
\label{sec:drl}

When solving a large scale multi-step 2p0s game with OSFP or SFP, 
it is challenging to directly solve SBR Eq. \eqref{eq:vi-sbr} due to the high dimension of $x \in \cX$, $y \in \cY$ where $K$ is large.
We propose to tackle it using Deep Reinforcement Learning (DRL).
The ingredients include: 
the function approximator for conditional policy, the policy gradient method and the opponent sampling.

% Take the viewpoint of Player 1 for the simplicity of discussion.
% Explain DRL: private Game Tree; information state (set); 
% counterfactual reach probability and reach probability
% define conditiaonal policy as the ratio of pi(a|s) = rho(s,a)/rho(s) = x(x,a)/x(s)
% say that we use function approximator to represent pi(a|s)

To begin with, 
we describe the (Single-Agent) Reinforcement Learning (RL).
For simplicity of the below discussion, 
let's take the perspective of Player 1.
RL aims for Player 1 to maximize the expected total reward in sequential decisions. 
This process can be organized as a \emph{Private Game Tree} consisting of \emph{nodes} and \emph{edges}.
Player 1 traverses the tree by starting at the root.
He can only take action at the node in hollow circle, 
called an \emph{information state} (or \emph{information set}) $s \in \cS$ where the set $\cS$ is all Player 1's information states and $|\cS| = K$ as in eq \eqref{eq:utility}.
Player 1 cannot take action at the node in solid circle where it is Player 2's turn to act.
In \emph{Imperfect Information Game} (IIG), 
an information state $s$ usually consists of multiple \emph{world states} (or simply \emph{states}), 
denoted by the set $\sigma(s)$ such that a world state $h \in \sigma(s)$.
At $s$ Player 1 has only \emph{partial observation} in the sense that he cannot distinguish from the world states $h \in \sigma(s)$.
For instance, in Strategy Card Game or Poker, Player 1 cannot see his opponent's hand cards,
where $s$ is his own hand cards and $h$ is the hand cards of both him and the opponent.
Therefore, at $s$ Player 1 has to take action $a$ from the action set associated to this information set $a \in \cA(s)$, 
disregarding what the world state $h \in \sigma(s)$ is under the hood.

For an \emph{edge} given by the pair $(s,a)$ and for the $x \in \cX$ as in Eq. \eqref{eq:utility}, 
denote by $x(s,a)$ the \emph{reach probability} for Player 1 to visit this edge when straining at the root node, 
excluding the opponent probability on the path.
Here the pair $(s,a)$ can also be understood as an integer index to some component of the vector $x$.
% denote by $x(s,a)$ the \emph{reach probability} for Player 1 to traverse this edge when straining at the root node.
Using Eq. \eqref{eq:utility} and \eqref{eq:payoff-vector}, 
the expected total reward (or \emph{Return}) for Player 1 is
\begin{equation}
    R \equiv -u(x, y) = \braket{x, -Ay} = \sum_{(s,a)} x(s,a) \cdot r(s,a),
\end{equation}
where the sum is taken over all Player 1 edges $(s,a)$ as $s \in \cS$, $a \in \cA(s)$, 
and $r(s,a)$ is the $(s,a)$-th component of the vector $-Ay$, 
which can be viewed as the \emph{immediate reward} associated with the edge $(s,a)$ and implicitly depends on Player 2 policy $y$.

For edge $(s,a)$ we define the full reach probability, 
taking into account both Player 1 policy $x$ and Player 2 policy $y$, 
as $\rho(s,a)$.
Define the reach probability for node $s$ as $\rho(s) = \sum_{a \in \cA(s)} \rho(s, a)$.
In a slight abuse of notation,
we let $x(s) \equiv \sum_{a \in \cA(s)} x(s,a)$ denote the reach probability for node $s$ contributed only by Player 1, 
which is also equal to $x(s',a')$ where traversing the edge $(s',a')$ immediately leads to $(s,a)$.
We thus define the \emph{conditional policy} as:
\begin{equation}
    \pi(a|s) \equiv \frac{\rho(s,a)}{\rho(s)} = \frac{x(s,a)}{x(s)} \quad s \in \cS, a \in \cA(s).
    \label{eq:cond-policy}
\end{equation}
In game theory literature, Eq. \eqref{eq:cond-policy} is called \emph{behavioral policy}; 
However, in RL literature, the term \emph{behavior policy} indicates the rollout policy in on-/off- policy learning. 
To avoid confusion, here we use the term ``conditional policy''.

To this extent,
we can represent $\pi(a|s)$ with a \emph{function approximator}.
In this work,
we parameterize the conditional policy as $\pi_{\theta}(a|s)$ using a neural network whose architecture is explained in Sec. \ref{sec:e2e-policy}.

The discussion goes similar for Player 2's Private Game Tree, 
and actually both Player 1's and 2's private tree can be derived from a \emph{Public Game Tree} \citep{schmid2021search}.

% Re-write SBR in normalized mixture form
Now let's get back to SBR as in eq \eqref{eq:osfp}. 
We continue to take the perspective of Player 1,
and henceforth focus on the expression relating to $\argmin_x$ only:
\begin{equation*}
    x_{k+1} = \argmin_{x \in \cX} \{ \eta \braket{\tilde{F}_k + F_k, x} + \psi(x) \},
\end{equation*}
where we slightly abuse notation by denoting the payoff vector as $F_i \equiv F(y_i) = A y_i$.
This is equivalent to solving 
\begin{equation}
    x_{k+1} = \argmin_{x \in \cX} \left \{ \sum_{i=1}^k \tilde{\alpha}_i \left ( \braket{F_i, x} + \frac{1}{\mu}\psi(x) \right ) \right \},
\end{equation}
where the objective function inside the $\argmin$ is scaled by a constant normalized factor given by:
\begin{equation*}
    \mu=\sum_{j=1}^k \eta \alpha_j, \quad \tilde{\alpha}_k = \frac{\eta \alpha_i}{\mu}, \quad \alpha_i = \begin{cases} 2 ,& i=k \\ 1 ,& i<k \end{cases}.
\end{equation*}
Let's rewrite the objective function as:
\begin{equation}
    \sum_{i=1}^k \tilde{\alpha}_i \ell_i(x),  \quad  \ell_i(x) \equiv \braket{F_i, x} + \frac{1}{\mu} \psi(x)
    \label{eq:sbr-obj}
\end{equation}
and take a closer look. 
Eq. \eqref{eq:sbr-obj} can be updated in a two-fold \emph{Stochastic Gradient Descent} (SGD):
The first is over sampling $i$, corresponding to opponent sampling;
The second is over $x$, corresponding to rollout in RL.

% Explain why (play against one model once a match) and how (sampling according to the weight) the opponent sampling
The gradient of the objective function Eq. \eqref{eq:sbr-obj} and be estimated as:
\begin{equation}
    \nabla_x \left ( \sum_{i=1}^k \tilde{\alpha}_i \ell_i(x) \right ) = \sum_{i=1}^k \tilde{\alpha}_i \nabla_x \ell_i(x) = E_{i \sim M(\tilde{\alpha}_i)} \left ( \nabla_x \ell_i(x) \right ),
\end{equation}
where $M(\tilde{\alpha}_i)$ is a multinomial distribution with weights $\tilde{\alpha}_i$ to ensure the estimate is unbiased.

% To sum-up PGT and gradient of \psi, say it can be learned by rollout segments.
Let's focus on $\ell_i(x)$ and rewrite it as 
\begin{equation}
    \ell_i(x) = R + \frac{1}{\mu}\psi, \quad R \equiv \braket{F_i, x}, \psi \equiv \psi(x).
    \label{eq:loss-i}
\end{equation}

For the first term in Eq. \eqref{eq:loss-i}, 
$R$ depends on policy $x \in \cX$, 
but $x$ is connected to the conditional policy $\pi(\cdot|\cdot)$ via eq \eqref{eq:cond-policy}. 
Thanks to the Policy Gradient Theorem \citep{sutton1999policy}, 
for an $s \in \cS$ and $a \in \cA(s)$, 
we are able to take the gradient wrt $\pi(a|s)$ as:
\begin{equation}
    \frac{\partial R}{\partial \pi(a|s)} = \rho(s) \cdot Q(s, a).
    \label{eq:pgt-iig}
\end{equation}
The policy gradient given in Eq. \eqref{eq:pgt-iig} is a reduced case which works on a tree structure,
while the original one \citep{sutton1999policy} is more general that covers the graph structure. 
Note also that Eq. \eqref{eq:pgt-iig} is in the context of IIG where $s \in \cS$ is information state instead of the world state $h$, 
therefore the state-action value $Q$ should be amended as \citep{zhang2021multi,srinivasan2018actor}:
\begin{equation}
    Q(s, a) = \frac{1}{\rho(s)} \cdot \sum_{\tau \in T(s,a)} \rho(\tau) R(\tau),
\end{equation}
where $\tau$ is a whole trajectory staring from root and terminating at a leaf,
$T(s, a)$ denotes the set of trajectories that traverse the edge $(s, a)$, 
$R(\tau)$ means the return of trajectory $\tau$ (e.g., sum of $r(s,a)$ for each edge along the path).

For the second term in Eq. \eqref{eq:loss-i}, 
the regularizer $\psi$ can be constructed as the weighted sum of the regularizer over the conditional policy for all information states \citep{farina2019optimistic}:
\begin{equation}
    \psi = \sum_{s \in \cS} x(s) \cdot \psi_{s}(\pi(\cdot|s)),
    \label{eq:regularizer-i}
\end{equation}
where $\psi_{s}(\cdot)$ can be identical and be simply negative entropy.

Combining Eq. \eqref{eq:pgt-iig} and \eqref{eq:regularizer-i}, 
we can solve \eqref{eq:loss-i} by standard RL techniques based on policy gradient method, 
e.g., PPO \citep{schulman2017proximal} or V-Trace \citep{espeholt2018impala}.
We can also adopt value bootstrap techniques (e.g., \citep{schulman2015high,espeholt2018impala}) to train on rollout segments instead of the whole trajectory.

The pseudo code of OSFP combined with DRL is given in Appendix \ref{app:osfp}.

% Remarks (put it here or move it to related work):
% * mixture of policies is in general NOT the mixture of conditional policies. citep Heinrich icml, citep sirinvenssa nips(or icml ?)
% * Two-fold SGD: one is sampling opponent, the other is sampling rollout 
% * last-iterate is more convenient to implement. cite Heinrich, cite Escher (poker, CFR stuff); say that we dont need a separate replay buffer for supervised learning
% * say our impl is (opponent) policy centric, while much other work is value centric. e.g., the (neural) CFR stuff (cite haobo paper)

% show the background of Strategy Card Game
\subsection{Strategy Card Game}
\label{sec:app_carg_game_bg}

\emph{Strategy Card Games} (SCGs) is a popular genre of modern strategy games. 
Classical SCGs include Magic: The Gathering, Gwent, Hearthstone, etc. 
%These games have complex rules and card types than board games, 
%and multi-stage and interconnected game flow with complicated decision logic. 
SCGs are considered more complicated than conventional board games such as chess and GO in several aspects:
vivid game rules, diversified card types and their interplay, imperfect information on the opponent, multi-stage decision making, to name a few.

SCGs usually have two stages: \emph{Card-Deck Building} (CB) and \emph{Battle} (BT). 
In the CB stage, 
both players build their own card deck by selecting a specified number of cards from the entire card pool,
then the game enters a multi-round BT stage. 
At the beginning of each BT round, 
players will gain mana related to the number of rounds, 
and draw a certain number of cards from the deck. 
Playing a card will consume the corresponding mana value. 
Players cannot play the card with a mana cost higher than the current mana value. 
The cards are played continuously until the end of the game. 

The effects of these cards can be roughly divided into two categories. 
The first category is the creature card, which generates creature units with specific attack, health, abilities and other attributes on an area (board) visible to both parties. 
These units can perform attacks and other operations. 
The specific operations that can be performed are related to the lane where the unit is located. 
The second category is the spell card, which equips players or units on the board some effects or additional skills. The outcome of a strategy card game is usually determined based on the health points (HP) of both parties, or the numerical attributes of the cards on the board.

\begin{figure}[t]
\centering
\includegraphics[width=1\textwidth]{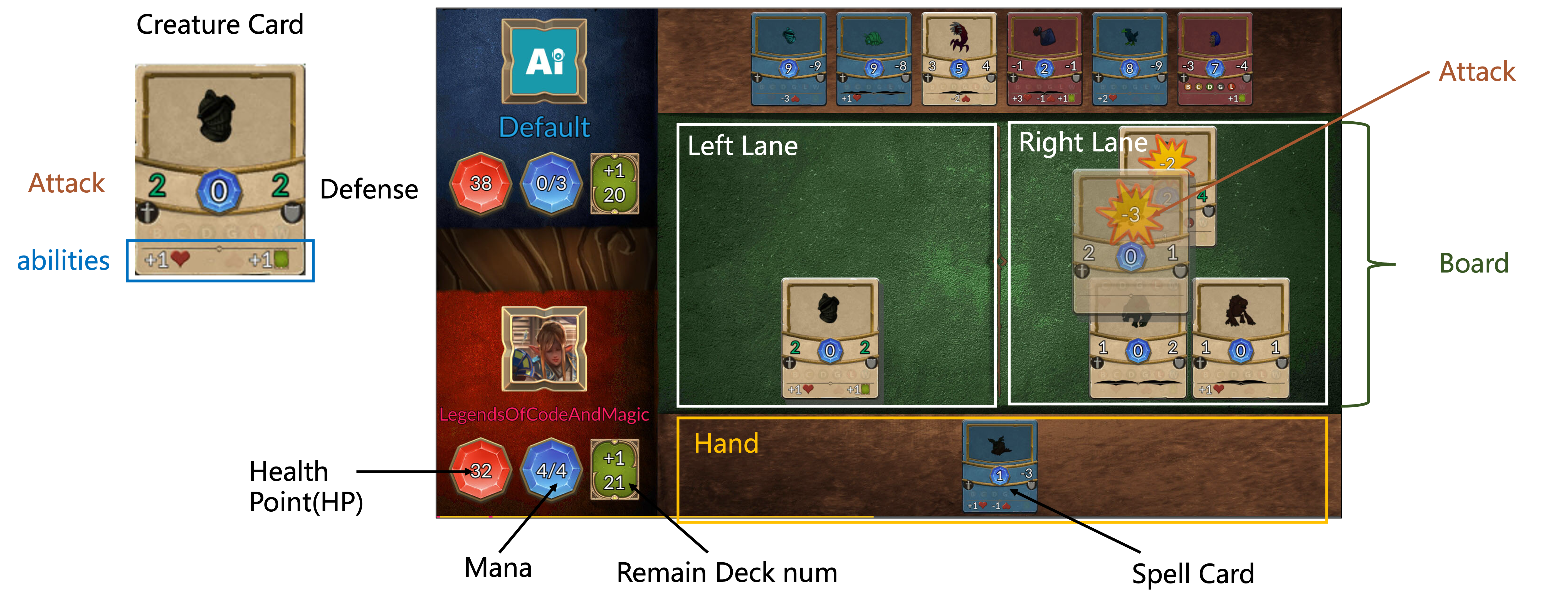}
\caption{LoCM game UI of BT stage. 
In the CB stage, the card deck is built by selecting either creature cards or spell cards, where each card has its own attributes. In the BT stage, the player has to keep own HP positive and try to hurt the opponent. 
Each card is played by consuming mana, which confines what cards can be played at each round. 
There are two lanes on which the player can choose to play a card.}
\label{fig:LoCM}
\end{figure}

\emph{Legends of Code and Magic} (LoCM),
developed by Jakub Kowalski and Radosław Miernik,
is a SCG specially designed for machine learning research (see the webpage \citep{locm}). 
It includes the basic SCG elements, 
and simplifies the effect and the representations of cards. 
We present LoCM game UI in Fig. \ref{fig:LoCM}. 
There are two game versions, LoCM 1.2 and LoCM 1.5, that are used by COG competitions \citep{locm}. 
There are two main differences between the two versions. 
One is that in LoCM 1.2, the card pool of CB stage contains 160 fixed cards. 
In the CB stage, 90 cards are randomly selected and divided into 30 rounds. In each round, one card is selected to add to the deck. 
In LoCM 1.5, the card pool of CB stage contains 120 randomly generated cards and 30 cards are selected to construct the deck, which requires a better understanding of the card attributes and the play strategy, and also increases the difficulty of the CB algorithm. 
The second is that the Area of Effect (AOE) is introduced in LoCM 1.5 so that some card is capable to affect all cards in the same lane or the both lanes.

% describe our network architecture and how it is trained with RL+Self-Play
\subsection{End-to-End Policy and RL Training}
\label{sec:e2e-policy}

To our best knowledge, 
previous submissions on LoCM 1.2 mostly use the following two-stage scheme. 
In the CB stage, 
deck is separately built with algorithms such as evolutionary algorithm\citep{yang2021deck}, Bayesian method, or combining expert rules and data mining to get some score function of cards. 
In the BT stage, 
the battle policy relies on search-based methods (e.g.,  \emph{Monte-Carlo Tree Search} (MCTS) \citep{browne2012survey}) or reinforcement learning \citep{vieira2019reinforcement}. 
We call this two-stage training scheme as \emph{Alternating Training} (AT). 
%This two-stage scheme has achieved good results in the previous competitions, but there exist some problems. 
Despite the good results shown in previous competitions, such a two-stage scheme can be problematic. 
%In the CB stage, it's difficult to deal with the randomness of the generated card pool in LoCM 1.5. 
In the CB stage, it's difficult to deal with the randomly generated card pool as in LoCM 1.5.
%It haven't built an explicit connection between the two stages during the training process in both LoCM 1.2 and LoCM 1.5.
Also, the metric for the deck quality is usually intuitive.
It is thus unclear how such a metric relates to the win-loss of the BT stage.

\begin{figure}[t]
\centering
\includegraphics[width=1\textwidth]{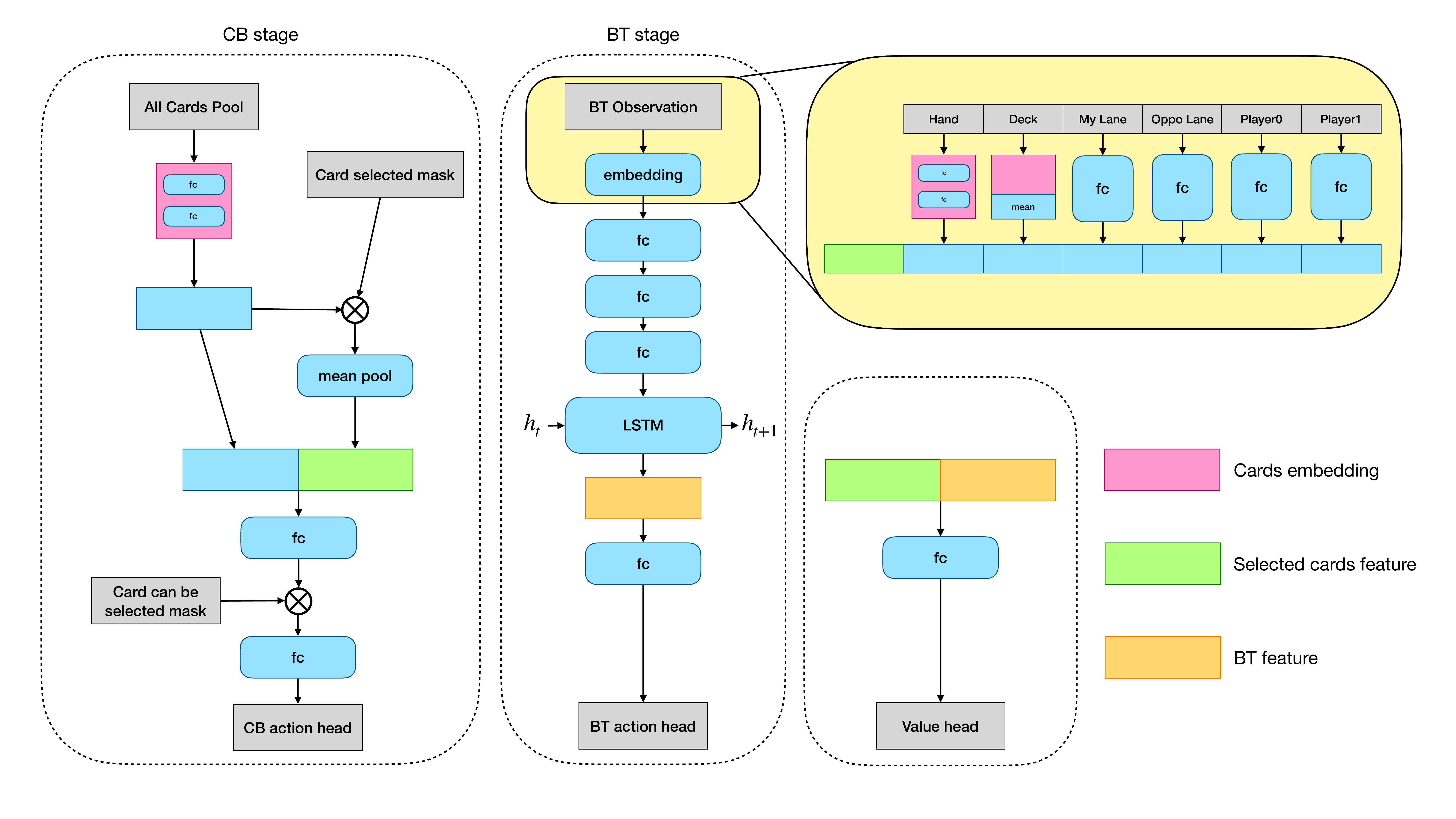}
\caption{E2E network architecture. \textit{Card selected mask} represents whether this card has been selected into the deck. \textit{Card can be selected mask} represents whether this card can be selected into the deck. }
\label{fig:network}
\end{figure}

To overcome the aforementioned drawbacks, we propose an E2E policy function, which trains the two stages of the game in a unified manner. We use a single representation function i.e. a single neural network to estimate CB policy, BT policy and value function. 
We treat each entire game as a trajectory, which contains both CB stage and BT stage, and utilize these trajectories to train our representation function end-to-end. 

We firstly setup the environment of LoCM. 
LoCM is a 2p0s IIG,
where the deck cards and hand cards of the opponent are not observable. 
When the opponent policy is stationary,
it reduces to \emph{Partially-Observable Markov Decision Process} (POMDP).
In the CB stage, 
the observation space (i.e., an information state) includes all candidate cards that can be selected and all selected cards of current deck for each step. 
The action space is a categorical distribution that selects one card from all candidate cards at each step. 
In LoCM 1.2, the CB stage contains 30 steps for each player, which is the same as the deck of 30 cards. All opponent's observations and actions are forbidden to be observed. 
This CB procedure is much like the \emph{Arena} mode in Hearthstone \citep{hearthstone-arena}.
In LoCM 1.5, the player is required to select 30 cards out from 120 randomly generated cards on each game episode beginning. 
We opt to split it into 30 steps, 
choosing one card once a time.
In this way, we deal with the CB stage in a unified procedure and henceforth with the same neural net architecture for both LoCM 1.2 and 1.5,
where a proper card-selection action mask is applied depending on the game version.
In the BT stage, 
the observation space includes all self hand cards, all self deck cards, cards on self and opponent lanes,
as well as scalar features for both players (e.g., HP, mana, no. of remaining deck cards). 
The action space is a single categorical distribution that contains all possible actions.
Like in CB stage, 
we also include an action mask zeroing the illegible actions at current time step depending on the state. 
We note that much of our observation/action space design is borrowed from the \emph{gym-locm project} \citep{gym-locm} developed by Ronaldo Vieira.
Finally, 
the reward is $+1$ at the end of the game if the player wins and is $-1$ otherwise. 

We use a single function approximator for the policy of both the CB and BT stage.  
Denote the observation (the information state) as $s \in \cS$ and the stage-indicatoras 
$$
\delta = \left\{\begin{aligned}1,\,\text{at CB stage},\\ 0,\,\text{at BT stage}. \end{aligned} \right.
$$
The output conditional policy is calculated by 
\begin{equation}
    \pi_\theta(\cdot|s) = \delta \cdot \pi_{\theta_{CB}}(\cdot|s) + (1 - \delta) \cdot \pi_{\theta_{BT}}(\cdot|s),
\end{equation}
where $\theta_{CB}$ and $\theta_{BT}$ are parameters of CB policy and BT policy respectively. 

The network architecture is shown in Fig. \ref{fig:network}.
Several remarks are in order.
i) The core module of CB policy is a deck cards embedding up to current time-step, 
which is obtained by mean-pooling the per-card embedding.
At CB stage, the deck cards embedding is updated recursively at each time-step (recall we select one card once a time),
while at BT stage the deck cards embedding gets fixed to provide the context information for BT policy.
ii) The per-card embedding layer is shared by the CB and BT policy, as the pink diagram in Fig. \ref{fig:network}.
In this way, the per-card feature representation is consistent for both the CB and BT stage.
%There exists overlapping parameters between $\theta_{CB}$ and $\theta_{BT}$, which is card embedding in our network, so the information between two stages can be shared and transmitted. 
iii) There is a unique value function that makes the prediction at every time-step in both the CB and BT stage,
and the input to this value function layer is the extracted feature from both the CB and BT policy.
Due to the BPTT of LSTM layer and the value bootstrapping mechanism in RL,
the reward signal at BT stage is able to ``penetrate backwards'' and impact the updating of the CB net parameter $\theta_{CB}$.

Previous two-stage scheme trains the CB policy and BT policy in an alternating way that the CB and BT stages are artificially isolated, which poses difficulties to the training signal transfer between the network sub-modules.
%and suffers from the hardness of information transmission between two isolated stages. 
In contrast, 
our E2E training scheme regards the two stages as a single episode of a POMDP,
which ensures the ``continuity'' of the reward signal along the entire trajectory. 
In addition, the parameter-sharing in CB/BT policy and the value function enhances the knowledge transfer among the network sub-modules. 

In the RL training,
we adopt policy gradient method where the value function is estimated by V-Trace \citep{espeholt2018impala}.
An auxiliary UPGO loss is also included \citep{vinyals2019grandmaster}. 
We employ negative entropy as the per information state regularizer $\psi_s(\cdot)$ as in Eq. \eqref{eq:regularizer-i}: 
\begin{equation}
\psi_s(p) = \sum_i - p_i \log p_i. 
\label{eq:entropy}
\end{equation}

\section{Related Work}
\label{sec:related_work}

DRL has made many impressive breakthroughs in a wide range of \emph{Markov Decision Processes} (MDP). 
From the perspective of planning, DRL has two categories, model-free methods \citep{ftw,vinyals2019grandmaster,badia2020agent57} and model-based methods \citep{silver2016mastering,schrittwieser2020mastering}. 
Model-free methods learn directly the interactions with the environment. 
When the dynamic of the environment is complicated, model-based methods learn the dynamic of the environment and plan on the learned dynamic model \citep{ha2018recurrent,okada2021dreaming}. 
When the dynamic of the environment is easy to simulate, model-based methods plan directly. 

From the perspective of learning algorithms, DRL has two categories, value-based methods \citep{dqn,hessel2018rainbow,r2d2} and  policy-based methods \citep{schulman2015trust,schulman2017proximal,espeholt2018impala}. 
Value-based methods learn state-action values and select the action according to their values. 
Policy-based methods learn a parameterized policy directly without consulting state-action values. 
Due to the complexity of the action space, it's usually difficult to estimate the state-action value of a complex game. Following \citep{vinyals2019grandmaster}, we train each action head independently by policy-based methods. 

DRL has two kinds of setups, offline \citep{fujimoto2019off,kumar2020conservative} and online. 
In this work, we concentrate on online setup. 
It's ideal to do on-policy training. However, due to the latent shift between the target policy and the behavior policy, the off-policy problem has to be dealt with. 
It's unbiased to do importance sampling over the entire trajectory but it has a high variance numerically \citep{sutton2018reinforcement}. 
Recent works \citep{retrace,espeholt2018impala} reduce the variance significantly by clipping per-step importance sampling. 

\citep{heinrich2015fictitious, heinrich2016deep} is the seminal work that combines Fictitious Play and (Deep) RL, 
but it only shows average-iterate convergence.
Henceforth, 
during RL training an extra replay buffer is adopted to collect rollouts from the historical models and the average policy is learned by a separate supervised learning. 

\citep{perolat2022mastering} also uses a single neural network to solve the two-stage game Stratego. 
The NE finding algorithm there is a discrete version of Replicator Dynamics, 
while in this work we use Optimistic Smooth Fictitious Play with standard single-agent policy gradient method as sub-solver.

\section{Experiments}
\label{sec:exp}

We firstly show our main result,
followed by the discussions on some topics.
%Then we introduce our work on some topics in the following parts. 

\subsection{Main Result}
\label{sec:main_result}

In the COG2022 competition, 
we won championships on both the ``official'' LoCM 1.5 track and the ``bonus'' LoCM 1.2 track. 
The average winrate against all other submissions on LoCM 1.5 is 84.41\%, 
and that on LoCM 1.2 is 94.56\%, 
as is shown in Fig. \ref{fig:reuslt} which are the results provided by the competition organizers. 

We carry out several internal evaluations using the code and models of each COG2022 submission 
that is released after the results announcement.
The head-to-head winrate on COG2022 LoCM 1.5 is shown in Table \ref{tab:result_1_5}, 
where each winrate is obtained by averaging $2500$ matches with random switching side.
Following the same criterion, 
we also evaluate our method in the ``Bonus'' LoCM 1.2 track (actually the submission from COG2021 LoCM 1.2 competition), 
and the results are show in Table \ref{tab:result_1_2} 
In both Table \ref{tab:result_1_5} and \ref{tab:result_1_2} the last column is the average winrate against all other submissions (i.e., the row-averaging).

\begin{table}[h]
\begin{center}
\scalebox{0.9}{
\begin{tabular}{c|c|c|c|c|c|c|c|c}
\toprule
winrate &ours	& NeteaseOPD	& Inspirai	& MSAODBV3	& USTC-gogogo	& Zylo	& RWI2L	& average \\
\midrule
ByteRL(ours) & n/a & 0.63 & 0.709 & 0.882 & 0.917 & 0.917 & 0.997 & 0.842 \\
NeteaseOPD & 0.37 & n/a & 0.575 & 0.839 & 0.874 & 0.83 & 0.99 & 0.746 \\
Inspirai & 0.291 & 0.425 & n/a & 0.804 & 0.78 & 0.762 & 0.984 & 0.674 \\
MSAODBV3 & 0.118 & 0.161 & 0.196 & n/a & 0.531 & 0.526 & 0.993 & 0.421\\
USTC-gogogo & 0.083 & 0.126 & 0.22 & 0.469 & n/a & 0.549 & 0.988 & 0.406 \\
Zylo & 0.083 & 0.17 & 0.238 & 0.474 & 0.451 & n/a & 0.972 & 0.398 \\
RWI2L & 0.003 & 0.01 & 0.016 & 0.007 & 0.012 & 0.028 & n/a & 0.013 \\
\bottomrule
\end{tabular}}
\caption{Evaluation on COG2022 LoCM 1.5. 
Submission ID abbreviations are 
MSAODBV3 for MugenSlayerAttackOnDuraraBallV3, 
RWI2L for RandomWItems2lanes.
}
\label{tab:result_1_5}
\end{center}
\end{table}

\begin{table}[h]
\begin{center}
\scalebox{0.7}{
\begin{tabular}{c|c|c|c|c|c|c|c|c|c|c|c|c|c}
\toprule
winrate &ours	& DrainPower	&DPA	&Chad	&Coac	&OLIE	&RG	&LANE\_1\_0	&Baseline1	&Baseline2	& AAA	&Ag2O	& average \\
\midrule
ByteRL(ours) &	n/a	&0.838	&0.846	&0.969	&0.875	&0.924	&0.968	&0.978	&1	&0.999	&0.997	&0.984	&0.943 \\ 
DrainPower	&0.162&	n/a	&0.512&	0.609&	0.634	&0.726	&0.755	&0.897	&0.966	&0.927&	0.94	&0.968&	0.736 \\ 
DPA	&0.154&	0.488&	n/a	&0.6	&0.63	&0.703	&0.743&	0.919	&0.964&	0.913&	0.923	&0.97&	0.728 \\ 
Chad	&0.031&	0.391&	0.4	&n/a	&0.439	&0.68&	0.98&	0.822&	0.954&	0.948	&0.954&	0.963&	0.687 \\ 
Coac	&0.125	&0.366&	0.37&	0.561&	n/a	&0.568&	0.661	&0.847&	0.972	&0.974	&0.945	&0.934&	0.666 \\ 
OLIE	&0.076	&0.274	&0.297	&0.32&	0.432&	n/a&	0.473&	0.832&	0.875&	0.912	&0.883	&0.745	&0.556 \\
RG	&0.032	&0.245	&0.257	&0.02	&0.339	&0.527&	n/a	&0.703	&0.957	&0.954	&0.934	&0.743	&0.519 \\ 
LANE\_1\_0	&0.022	&0.103	&0.081	&0.178	&0.153	&0.168	&0.297	&n/a	&0.658	&0.635	&0.71	&0.755	&0.342 \\ 
Baseline1	&0	&0.034	&0.036	&0.046	&0.028	&0.125	&0.043	&0.342	&n/a	&0.4	&0.764	&0.611	&0.221 \\
Baseline2&	0.001&	0.073	&0.087	&0.052	&0.026	&0.088	&0.046	&0.365	&0.6	&n/a	&0.298	&0.611	&0.204 \\ 
AAA	&0.003	&0.06	&0.077	&0.046	&0.055	&0.117	&0.066	&0.29	&0.236	&0.702	&n/a	&0.561	&0.201 \\ 
Ag2O	&0.016	&0.032	&0.03	&0.037	&0.066	&0.255	&0.257	&0.245	&0.389	&0.389	&0.439	&n/a	&0.196 \\
\bottomrule
\end{tabular}}
\caption{Evaluation on COG2021 LoCM 1.2. 
Submission ID abbreviations are 
AAA for AdvancedAvocadoAgent,
DPA for DrainPowerAgg,
OLIE for OneLaneIsEnough,
RG for ReinforcedGreediness.
}
\label{tab:result_1_2}
\end{center}
\end{table}

For the Fictitious Play and DRL training, 
we use an internal framework that is much like the open-sourced library TLeague \citep{sun2020tleague}.
Both GPUs (V100) and CPU cores are employed in a single training session, 
where we fix the ratio as ``1 GPU drags 600 CPU cores'' and refer to it as ``1 Unit''.
The submitted model for COG2022 LoCM 1.5 was trained with ``24 Units'' for around 72 hours, 
where the throughput during training is approximately 500K fps (i.e., 500K observations are produced and consumed as training samples per second).
The submitted model for LoCM 1.2 was firstly trained with ``1 Unit'' for 3 days and then continued training with ``8 Units'' for 6 days.

\subsection{Card Embedding Sharing}
\label{sec:card_embed_share}

We performed several ablation experiments to test the neural network structure.
The criterion is to perform matches of ``model-with-some-feature'' vs ``model-without-this-feature'' and see how much the winrate exceeds $50\%$.
Like the aforementioned internal evaluation, 
we run 2500 matches with random switching side.
We found that the shared selected cards feature (see Fig. \ref{fig:network}) obtains a winrate $53\%$, 
and the shared cards embedding (see Fig. \ref{fig:network}) obtains a winrate $55\%$. 
Hence, we always use shared parameters in our experiments, except for Alternating Training. 
For ablation study in Sec. \ref{sec:alt_train}, all approaches stop sharing parameters for fairness. 

\subsection{Alternating Training}
\label{sec:alt_train}

\begin{table}[h]
\begin{center}
\scalebox{0.9}{
\begin{tabular}{cc|cccc}
\toprule
Version & Baseline & Evo-AT & Neural-AT & E2E \\
\midrule
LoCM 1.2 & DrainPower & 64.7\% & 68\% & 81\% \\
LoCM 1.5 & Evo-AT & 50\% & 56.5\% & 65.5\% \\
\bottomrule
\end{tabular}}
\caption{Alternating Training. Each entry represents the winrate of the column scheme vs the row baseline. All experiments are within the same resource budget except for DrainPower. }
\label{tab:at}
\end{center}
\end{table}

We have also implemented AT and compared with our E2E scheme. 
Each learning period of AT only trains one stage of the game. The stage switching is based on the winrate and the number of learning periods. 
The AT training process is shown in Appendix \ref{app:at}. 
We have tried two AT schemes. The difference between the two is the CB stage. One exploits expert priors to artificially construct the cost-efficiency curves 
% (The relationship between card effects and mana consumption which are determined by multiple artificially defined parameters), 
and apply an evolutionary algorithm to update the CB policy. 
The other directly applies reinforcement learning to obtain the CB policy. 
For convenience, we denote them as Evo-AT and Neural-AT respectively. 
% In order to reduce the impact of the network structure on the comparison, the alternately trained network structure used in this paper only removes information sharing on the end-to-end network structure.
The result is shown in Table \ref{tab:at} and our E2E scheme shows the best performance.

\subsection{Post-Process}
\label{sec:post}

One post-process technique is to adjust the temperature when evaluating the model, which is useful in our model.  
Denote the logits function of $\pi(\cdot|s)$ as $l(\cdot|s)$ for some information state $s$. 
During the training process, we sample the action by $a \sim \text{softmax}(l(\cdot|s))$. 
When we evaluate the model, it's capable to add one temperature $\tau$ to sample action by $a \sim \text{softmax}(l(\cdot|s) / \tau)$. 
Note that $\tau = 0+$ is equivalent to $\text{argmax}$ as $\text{softmax}(\cdot / \tau)|_{\tau = 0+} = \text{argmax}(\cdot)$. 
On our one day model, we have verified that $\tau = 0+$ shows a better performance, which has $53\%$ winrate vs $\tau = 1.0$. 
Throughout all our experiments, we always use $\text{argmax}$ to do evaluation. 

The other post-process technique is One Turn Kill (OTK). 
After COG2022 LoCM 1.5, we noticed that team NeteaseOPD applies an OTK technique. 
We removed OTK from NeteaseOPD, re-ran the competition and found the average winrate of NeteaseOPD dropped $2\%$. 
However, in our submission, adding OTK rule seems not to significantly change the winrate, 
an evidence that our model has learned a precise calculation on the HP number when delivering efficient attacking.

\subsection{Monte-Carlo Tree Search}
\label{sec:mcts}

Post-process techniques are mostly rule-based and may introduce undesired bias to the policy. 
When looking for better substitution, it's natural to consider planning on this card game. 
On LoCM 1.2, 
we have also implemented Information State Monte-Carlo Tree Search (IS-MCTS).
Despite IS-MCTS is game theoretically ``unsafe'' as it neglects the entering probability at an information state due to the opponent policy (cf \citep{brown2020combining} for explanations), 
it indeed shows good performance on some card games \citep{cowling2012information}.
Since LoCM is a POMDP, 
where the opponent's hand and deck are not observable, 
instead of guessing a world state from the feasible world state set (i.e., all possible opponent hands and decks at an information state), 
we simply use the ground truth world state as the guessing world state to expand and search. 
In the following we simply call IS-MCTS as MCTS for short without confusion. 
We present our detailed implementation of MCTS in Appendix \ref{app:mcts}. 

We apply MCTS to sample actions and compare MCTS with pure model-free training within the same computational resource budget and the same wall-time. 
We use $(p, n, m)$ to denote the parameter of MCTS, where $p$ represents the probability whether we expand MCTS on each information state, $n$ is the number of expansions on that information state, and $m$ is the number of successive information states to expand. 
For instance, at timestamp $i$, if we have $\text{Unif}(0, 1) < p$, then we expand $n$ nodes on every state of timestamp $i, \dots, i+m-1$. 
The result is shown in Table \ref{tab:mcts}. 

\begin{table}[h]
\begin{center}
\scalebox{0.9}{
\begin{tabular}{c|c|c|c|c|c}
\toprule
 MCTS parameters & (0.1, 40, 1)	& (0.02, 200, 1)	& (0.002, 200, 10)	& (0.001, 200, 20)	& (0.00025, 400, 40) \\
\midrule
winrate (vs baseline) & 36\% & 46\% & 46\% & 50\% & 56\% \\
\bottomrule
\end{tabular}}
\caption{MCTS with different parameters. Baseline (no MCTS) is trained with 1 GPU for 40 hours. All MCTS experiments are trained with the same resource budget as baseline. }
\label{tab:mcts}
\end{center}
\end{table}

\begin{table}[h]
\begin{center}
\scalebox{1.0}{
\begin{tabular}{c|c|c|c}
\toprule
 training time & 40h & 80h & 144h \\
\midrule
(0.00025, 400, 40) vs baseline & 56\% & 52\% & 51\% \\
\bottomrule
\end{tabular}}
\caption{MCTS with different training time. Comparisons are under the same resource budget and the same physical training time. }
\label{tab:mcts_time}
\end{center}
\end{table}

Since MCTS reduces the speed of data generation, $(0.1, 40, 1)$ drops a lot on winrate within the same resource budget. 
But when we apply MCTS to produce higher quality samples by expanding more on a single state and expanding on more successive states, we find the performance increases monotonically without change of expected number of expansions. 
However, this enhancement is inconsistent with respect to wall-time: when we continue training, we find the advantage of MCTS diminishes, which is shown in Table \ref{tab:mcts_time}. 
Therefore, we haven't applied MCTS, because it produces less samples under the same resource budget and its promotion is unsatisfactory. 

\section{Ethics Statement}

This work is a pure academic study in deep reinforcement learning on strategy card games. It may provide opportunities for malicious applications in the future, but we do not encourage and will not support any dishonorable activities.

\section{Conclusions}
\label{sed:conclusion}

In this work, we study a strategy card game and win double championship in COG2022 competitions. 
We propose an end-to-end policy function to deal with the two-stage game. 
We propose an optimistic smooth fictitious play strategy to find the Nash equilibrium of the two-player zero-sum game. 
We also do extensive studies to verify the advancement of our approach and make a discussion about related topics. 
We hope our work could establish a standard solution for strategy card games. 

\bibliographystyle{iclr2023}
\bibliography{refs}

\input{appendix-arxiv20230308.tex}

\end{document}

%% file: appendix-arxiv20230308.tex
\clearpage

\appendix

\section{Hyperparameters}
\label{app:hp}

\begin{table}[h]
    \centering
    \begin{tabular}{c|c}
        \toprule
         Parameter & Value \\
         \midrule
         Weight of policy gradient from V-Trace & 1.0 \\
         Weight of policy gradient from UPGO & 1.0 \\
         Weight of value function loss & 1.0 \\
         Weight of entropy penalty & 0.01 \\
         Learning rate & 5e-5 \\
         Batch size & 4e+4 * no. GPUs \\
         Discount & 0.99 \\
         LSTM states & 256 \\
         Sample reuse & 2 \\
         V-Trace $c$ clip & 1.0 \\
         V-Trace $\rho$ clip & 1.0 \\
         \bottomrule
    \end{tabular}
    \caption{Reinforcement learning hyperparameters. }
    \label{tab:rl_hyper_parameter}
\end{table}

\begin{table}[h]
    \centering
    \begin{tabular}{c|c}
        \toprule
         Parameter & Value \\
         \midrule
         Self-play Probability, $p$ & 0.6 \\
         Add to historical model threshold, $\xi$ & 0.7 \\
         Add to historical model max LP, $c$ & 6 \\
         Num samples of each LP & 8e+8 \\
         \bottomrule
    \end{tabular}
    \caption{OSFP hyperparameters. The notions are corresponding to Alg. \ref{alg:ofsp}.}
    \label{tab:osfp_hyper_parameter}
\end{table}

% \clearpage
\section{Optimistic Smooth Fictitious Play}
\label{app:osfp}

% \begin{wraptable}[24]{r}{2.9in}
%   \vspace{-0.3cm}
\begin{center}
\begin{minipage}{0.5\linewidth}
\begin{algorithm}[H]
\caption{Optimistic Smooth Fictitious Play. }  
  \begin{algorithmic}
    \STATE Init $H = []$.
    \STATE Init $0 < p < 1, 0 < \xi < 1$. 
    \STATE Init $c > 0$, $count = 0$.
    \STATE Init the probability function $f$. 
    \FOR{$LP=0,1,2,\dots$}
        \STATE Init $G = C = [0] \text{ * len}(H)$. 
        \WHILE{LP not end}
            \FOR{each actor}
                \IF{len($H$) = 0 \OR $\text{Unif}(0, 1) < p$}
                    \STATE Opponent player = current learner.
                    \STATE Finish game. 
                \ELSE
                    % \STATE Sample $i \sim \frac{(1 - G[i] / C[i])^2}{\sum_i (1 - G[i] / C[i])^2}$. 
                    \STATE Sample $i \sim f(\dots, (G[i], C[i]), \dots)$.
                    \STATE Opponent Player = $H[i]$.
                    \STATE Finish game and get $g \in \{+1, -1\}$.
                    \STATE $G[i] = G[i] + g$, $C[i] = C[i] + 1$. 
                \ENDIF
            \ENDFOR
        \ENDWHILE
        \IF{($G[i] / C[i] > \xi, \forall\,i$) \OR ($count > c$)}
            \STATE Add current learner to $H$. 
            \STATE $count = 0$. 
        \ELSE
            \STATE $count = count + 1$. 
        \ENDIF
    \ENDFOR
  \end{algorithmic}
\label{alg:ofsp}
\end{algorithm}
\end{minipage}
\end{center}
% \end{wraptable}

\section{Alternating Training}
\label{app:at}

The training schedule of alternating training is shown below. The adjustments are highlighted in blue. 

\begin{center}
\begin{minipage}{0.5\linewidth}
\begin{algorithm}[H]
\caption{Alternating Training. }  
\begin{algorithmic}
\STATE Init $H = []$.
\STATE Init $0 < p < 1, 0 < \xi < 1$. 
\STATE Init $c > 0$, $count = 0$.
\STATE Init the probability function $f$. 
\STATE {\color{blue} Init training stage s = BT.} 
\FOR{$LP=0,1,2,\dots$}
    \STATE Init $G = C = [0] \text{ * len}(H)$. 
    \WHILE{LP not end}
        \FOR{each actor}
            \IF{len($H$) = 0 \OR $\text{Unif}(0, 1) < p$}
                \STATE Opponent player = current learner.
                \STATE Finish game. 
            \ELSE
                % \STATE Sample $i \sim \frac{(1 - G[i] / C[i])^2}{\sum_i (1 - G[i] / C[i])^2}$.
                \STATE Sample $i \sim f(\dots, (G[i], C[i]), \dots)$.
                \STATE Opponent Player = $H[i]$.
                \STATE Finish game and get $g \in \{+1, -1\}$.
                \STATE $G[i] = G[i] + g$, $C[i] = C[i] + 1$. 
            \ENDIF
        \ENDFOR
    \ENDWHILE
    \IF{($G[i] / C[i] > \xi, \forall\,i$) \OR ($count > c$)}
        \STATE Add current learner to $H$. 
        \STATE $count = 0$. 
        \STATE {\color{blue}Change stage s = \{CB, BT\} - s.}
    \ELSE
        \STATE $count = count + 1$. 
    \ENDIF
\ENDFOR
\end{algorithmic}
\label{alg:at}
\end{algorithm}
\end{minipage}
\end{center}

% \clearpage

\section{Monte-Carlo Tree Search}
\label{app:mcts}

Since LOCM is a POMDP, where the opponent's hand and deck are not observable, 
we cannot directly apply planning as in Perfect Information Game. 
An alternative is to sample the world states (e.g., sample opponent hand and deck) from an information state and take expectation over them \citep{cowling2012information}. 
Due to the complexity and expansive computations, 
we adopt a simple way and only expand the tree on the ground truth world state. 

Since our E2E policy $\pi$ can be applied directly on both CB stage and BT stage, 
we do not distinguish the two stages and always write $\pi(\cdot|s)$ as the conditional policy for an information state $s \in \cS$. 
%We denote the information state (partial observation) as $o$. 
We write the value estimation as $v(s)$. 
Thanks to the E2E policy, 
MCTS is capable to start at a root world state at any stage (CB or BT) and to end on a leaf at any stage later. 
Without loss of generality, 
we assume that the root world state is in player 1's turn. 
When the expansion meets player 2's turn, 
since there exist multiple steps in a turn, 
we repeat sampling by the conditional policy $\pi(\cdot|s)$ (over player 2's information state $s$) until that player 2's turn finishes and continue expanding player 1's tree.
where  
After MCTS, we use the normalized visiting counts as the probability to sample the action of current root information state. 
We also use this probability as the behavior policy whenever importance sampling is required in RL. 

Formally, we provide notations and describe the entire process of expanding on a root information state in Alg. \ref{alg:mcts}. 
Denote the world state as $h$, 
the corresponding information states (partial observations) for player 1 and player 2 as $s_1(h), s_2(h)$ respectively. 
Notice that the dynamic of the environment is a transition $h \rightarrow h'$, 
but the conditional policy observes partially to act as $\pi(\cdot|s(h))$. 
Denote the action profile of player 1 and 2 as a tuple, 
and denote the dynamic of the environment as $h' = envstep(h, (a, \varnothing)) = envstep(h, (s_1(h), \varnothing))$ in player 1's turn and $h' = envstep(h, (\varnothing, a)) = envstep(h, (\varnothing, s_2(h)))$ in player 2's turn. 
We use $\varnothing$ to represent \emph{empty}. 
Denote the node of the tree as $N$, 
where $N.child$ represents child nodes, 
$N.parent$ represents the parent node, 
$N.h$ represents the world state, 
$N.a$ represents the action from the parent node to current node, 
$N.p$ represents the probability from the parent to current node, 
$N.n$ represents visitation count, 
$N.w$ represents the summation of values and $N.q$ represents the world state-action value of current node.

% \begin{wraptable}[28]{m}{4.0in}
% \vspace{-0.5cm}
\centering
\begin{minipage}{0.62\linewidth}
\begin{algorithm}[H]
\caption{Perfect State Monte-Carlo Tree Search. }  
\begin{algorithmic}
\STATE Init $n=400$, $c=5.0$, $\tau = 10.0$, $\alpha=0.03$, $p=0.75$. 
\STATE Init root state $h_0$, root node $N_0$.
\STATE Let $N_0.h$ = $h_0$, $N_0.parent = \varnothing$, $N_0.child = \varnothing$.
\FOR{i = 1, 2, \dots, n}
    \STATE 
    \STATE // SELECT
    \STATE Let $N = N_0$. 
    \WHILE{$N.child$ is not $\varnothing$}
        \STATE $N = {\arg\max}_{N \in N.child} N.q + c \cdot \frac{\sqrt{1 + \sum_{N \in N.child} N.n}}{1 + N.n} \cdot N.p$.
    \ENDWHILE
    \STATE 
    \STATE // EXPAND
    \STATE $h = envstep(N.parent.h, (N.a, \varnothing))$. 
    {\color{blue}\WHILE{$h$ is in player 2's turn}
        \STATE $h = envstep(h, (\varnothing, \pi(\cdot|s_2(h))))$.
    \ENDWHILE}
    \STATE Let $N.h = h$. 
    \STATE $x = softmax(\log \pi(\cdot|s_1(h)) / \tau)$, $d = Direchlet(\alpha)$. 
    \FOR{$a$ in available actions of $h$}
        \STATE Init node $\Tilde{N}$.
        \STATE Let $\Tilde{N}.parent = N$, $\Tilde{N}.child = \varnothing$, $\Tilde{N}.n = \Tilde{N}.w = 0$, $\Tilde{N}.a = a$, $\Tilde{N}.p = p \cdot x[a] + (1 - p) \cdot d[a]$. 
        \STATE Add $\Tilde{N}$ to $N.child$. 
    \ENDFOR
    \STATE 
    \STATE // BACKTRACE
    \STATE $v = v(s_1(h))$. 
    \WHILE{$N$ is not $n/a$}
        \STATE $N.n = N.n + 1$, $N.w = N.w + v$. 
        \STATE $N.q = N.w / N.n$. 
        \STATE $N = N.parent$. 
    \ENDWHILE
\ENDFOR
\STATE 
\STATE Let $\pi_{MCTS}(\cdot|s_1(h_0)) = \frac{1}{\sum_{N \in N_0.child} N.n} (\dots, N.n, \dots)$. 
\STATE Sample $a \sim \pi_{MCTS}(\cdot|s_1(h_0))$. 
\RETURN $a, \pi_{MCTS}(\cdot|s_1(h_0))$, $N_0.q$.  
\end{algorithmic}
\label{alg:mcts}
\end{algorithm}
\end{minipage}
% \end{wraptable}